\documentclass{article}

\usepackage[final]{nips_2018}

\usepackage[utf8]{inputenc} %
\usepackage[T1]{fontenc}    %
\usepackage{hyperref}       %
\usepackage{url}            %
\usepackage{booktabs}       %
\usepackage{amsfonts}       %
\usepackage{nicefrac}       %
\usepackage{microtype}      %

\usepackage{graphicx}
\usepackage{amsmath}
\usepackage{amssymb}
\usepackage{color}
\usepackage{amsthm}
\usepackage{placeins}
\usepackage{wrapfig}
\usepackage{subcaption}
\usepackage{nicefrac}

\usepackage{algorithm}
\usepackage{algorithmicx}  
\usepackage{algpseudocode}

\usepackage{hyperref}

\renewcommand{\cite}{\citep}

\newtheorem{theorem}{Theorem}
\newtheorem{lemma}{Lemma}

\newcommand{\eg}{e.g.\ }

\newcommand{\E}[2]{\operatorname{\mathbb{E}}_{#1}\left[#2\right]}

\newcommand{\density}{p}
\newcommand{\kl}[2]{\mathrm{D_{KL}}\left(#1\;\middle\|\;#2\right)}
\newcommand{\entropy}{\mathcal{H}}
\newcommand{\ent}{\mathcal{H}}

\newcommand{\voidarg}{{\,\cdot\,}}

\newcommand{\sspace}{\mathcal{S}}
\newcommand{\aspace}{\mathcal{A}}

\newcommand{\state}{\mathbf{s}}

\newcommand{\st}{{\state_t}}

\newcommand{\sT}{{\state_T}}
\newcommand{\stp}{{\state_{t+1}}}

\newcommand{\pdyn}{\density}

\newcommand{\action}{\mathbf{a}}

\newcommand{\at}{{\action_t}}

\newcommand{\atp}{{\action_{t+1}}}
\newcommand{\aT}{{\action_T}}

\newcommand{\opt}{^*}

\newcommand{\urv}{\mathbf{u}}

\newcommand{\reward}{r}

\newcommand{\rmin}{r_\mathrm{min}}
\newcommand{\rmax}{r_\mathrm{max}}

\newcommand{\V}{V}

\newcommand{\Q}{Q}

\newcommand{\policy}{\pi}
\newcommand{\policyold}{{\policy_\mathrm{old}}}
\newcommand{\policynew}{{\policy_\mathrm{new}}}

\newcommand{\policyopt}{{\policy^*}}

\newcommand{\params}{\theta}

\newcommand{\pparams}{{\phi}}   % policy parameters
\newcommand{\gauss}{\mathcal{N}}

\newcommand{\reals}{\mathbb{R}}

\newcommand{\discount}{\gamma}

\newcommand{\aref}[1]{\hyperref[#1]{Appendix~\ref*{#1}}}

\title{Soft Actor-Critic Algorithms and Applications}

\newcommand{\equal}{$^*$}
\newcommand{\ucb}{$^\dagger$}
\newcommand{\brain}{$^\ddagger$}

\author{
Tuomas Haarnoja\equal\ucb\brain
\And
Aurick Zhou\equal\ucb
\And
Kristian Hartikainen\equal\ucb
\And
George Tucker\brain
\And
Sehoon Ha\brain
\And
Jie Tan\brain
\And
Vikash Kumar\brain
\And
Henry Zhu\ucb
\And 
Abhishek Gupta\ucb
\And
Pieter Abbeel\ucb
\And
Sergey Levine\ucb\brain
}

\begin{document}

\newcommand\blfootnote[1]{%
  \begingroup
  \renewcommand\thefootnote{}\footnote{#1}%
  \addtocounter{footnote}{-1}%
  \endgroup
}
\blfootnote{\ucb UC Berkeley, \brain Google Brain, \equal Contributed equally}

\maketitle

\begin{abstract}
Model-free deep reinforcement learning (RL) algorithms have been successfully applied to a range of challenging sequential decision making and control tasks. However, these methods typically suffer from two major challenges: high sample complexity and brittleness to hyperparameters. Both of these challenges limit the applicability of such methods to real-world domains. In this paper, we describe Soft Actor-Critic (SAC), our recently introduced off-policy actor-critic algorithm based on the maximum entropy RL framework. In this framework, the actor aims to simultaneously maximize expected return and entropy; that is, to succeed at the task while acting as randomly as possible. We extend SAC to incorporate a number of modifications that accelerate training and improve stability with respect to the hyperparameters, including a constrained formulation that automatically tunes the temperature hyperparameter. We systematically evaluate SAC on a range of benchmark tasks, as well as challenging real-world  tasks such as locomotion for a quadrupedal robot and robotic manipulation with a dexterous hand. With these improvements, SAC achieves state-of-the-art performance, outperforming prior on-policy and off-policy methods in sample-efficiency and asymptotic performance. Furthermore, we demonstrate that, in contrast to other off-policy algorithms, our approach is very stable, achieving similar performance across different random seeds. These results suggest that SAC is a promising candidate for learning in real-world robotics tasks.
\end{abstract}

\section{Introduction}

Model-free deep reinforcement learning (RL) algorithms have been applied in a range of challenging domains, from games~\citep{mnih2013playing,silver2016mastering} to robotic control~\citep{gu2017deeprobot,haarnoja2018composable}. The combination of RL and high-capacity function approximators such as neural networks holds the promise of automating a wide range of decision making and control tasks, but widespread adoption of these methods in real-world domains has been hampered by two major challenges. First, model-free deep RL methods are notoriously expensive in terms of their sample complexity. Even relatively simple tasks can require millions of steps of data collection, and complex behaviors with high-dimensional observations might need substantially more. Second, these methods are often brittle with respect to their hyperparameters: learning rates, exploration constants, and other settings must be set carefully for different problem settings to achieve good results. Both of these challenges severely limit the applicability of model-free deep RL to real-world tasks.

One cause for the poor sample efficiency of deep RL methods is on-policy learning: some of the most commonly used deep RL algorithms, such as TRPO~\cite{schulman2015trust}, PPO~\citep{schulman2017proximal} or A3C~\citep{mnih2016asynchronous}, require new samples to be collected for (nearly) every update to the policy. This quickly becomes extravagantly expensive, as the number of gradient steps and samples per step needed to learn an effective policy increases with task complexity. Off-policy algorithms aim to reuse past experience. This is not directly feasible with conventional policy gradient formulations, but is relatively straightforward for Q-learning based methods~\citep{mnih2015human}. Unfortunately, the combination of off-policy learning and high-dimensional, nonlinear function approximation with neural networks presents a major challenge for stability and convergence~\citep{bhatnagar2009convergent}.
This challenge is further exacerbated in continuous state and action spaces, where a separate actor network is often used to perform the maximization in Q-learning. 

In~\citep{haarnoja2018soft}, we introduced the Soft Actor-Critic (SAC) algorithm based on the maximum entropy framework~\citep{ziebart2008maximum,toussaint2009robot,rawlik2012stochastic,fox2015taming,haarnoja2017reinforcement}. In the first sections of this paper, we summarize the SAC algorithm, describe the reasoning behind the design choices, and present key theoretical results from~\citep{haarnoja2018soft}. Unfortunately, SAC as presented in~\citep{haarnoja2018soft} can suffer from brittleness to the temperature hyperparameter. Unlike in conventional reinforcement learning, where the optimal policy is independent of scaling of the reward function, in maximum entropy reinforcement learning the scaling factor has to be compensated by the choice a of suitable temperature, and a sub-optimal temperature can drastically degrade performance~\cite{haarnoja2018soft}. To resolve this issue, we devise an  automatic gradient-based temperature tuning method that adjusts the expected entropy over the visited states to match a target value. Although this modification is technically simple, we find that in practice it largely eliminates the need for per-task hyperparameter tuning. %
Finally, we present empirical results that show that Soft Actor-Critic attains a substantial improvement in both performance and sample efficiency over prior off-policy and on-policy methods including the recently introduced twin delayed deep deterministic (TD3) policy gradient algorithm~\citep{fujimoto2018addressing}. We also evaluate our method on real-world challenging tasks such as locomotion for a quadrupedal robot and robotic manipulation with a dexterous hand from image observations.

\section{Related Work}
Maximum entropy reinforcement learning generalizes the expected return RL objective, although the original objective can be recovered in the zero temperature limit~\citep{haarnoja2017reinforcement}.
More importantly, the maximum entropy formulation provides a substantial improvement in exploration and robustness: as discussed by \citet{ziebart2010modeling}, maximum entropy policies are robust in the face of model and estimation errors, and as demonstrated by~\cite{haarnoja2017reinforcement}, they improve exploration by acquiring diverse behaviors. Prior work has proposed model-free deep RL algorithms that perform on-policy learning with entropy maximization~\citep{o2016pgq}, as well as off-policy methods based on soft Q-learning and its variants~\citep{schulman2017equivalence, nachum2017bridging, haarnoja2017reinforcement}. However, the on-policy variants suffer from poor sample complexity for the reasons discussed above, while the off-policy variants require complex approximate inference procedures in continuous action spaces.

Our soft actor-critic algorithm incorporates three key ingredients: an actor-critic architecture with separate policy and value function networks, an off-policy formulation that enables reuse of previously collected data for efficiency, and entropy maximization to encourage stability and exploration. We review prior work that draw on some of these ideas in this section. Actor-critic algorithms are typically derived starting from policy iteration, which alternates between \emph{policy evaluation}---computing the value function for a policy---and \emph{policy improvement}---using the value function to obtain a better policy~\citep{barto1983neuronlike,sutton1998reinforcement}. In large-scale reinforcement learning problems, it is typically impractical to run either of these steps to convergence, and instead the value function and policy are optimized jointly. In this case, the policy is referred to as the actor, and the value function as the critic. Many actor-critic algorithms build on the standard, on-policy policy gradient formulation to update the actor~\citep{peters2008reinforcement}, and many of them also consider the entropy of the policy, but instead of maximizing the entropy, they use it as an regularizer \citep{schulman2017proximal,schulman2015trust,mnih2016asynchronous,gruslys2017reactor}. On-policy training tends to improve stability but results in poor sample complexity.

There have been efforts to increase the sample efficiency while retaining robustness by incorporating off-policy samples and by using higher order variance reduction techniques~\citep{o2016pgq,gu2016q}. However, fully off-policy algorithms still attain better efficiency. A particularly popular off-policy actor-critic method, DDPG~\cite{lillicrap2015continuous}, which is a deep variant of the deterministic policy gradient~\citep{silver2014deterministic} algorithm, uses a Q-function estimator to enable off-policy learning, and a deterministic actor that maximizes this Q-function. As such, this method can be viewed both as a deterministic actor-critic algorithm and an approximate Q-learning algorithm. Unfortunately, the interplay between the deterministic actor network and the Q-function typically makes DDPG extremely difficult to stabilize and brittle to hyperparameter settings~\citep{duan2016benchmarking,henderson2017deep}. As a consequence, it is difficult to extend DDPG to complex, high-dimensional tasks, and on-policy policy gradient methods still tend to produce the best results in such settings~\citep{gu2016q}. Our method instead combines off-policy actor-critic training with a \emph{stochastic} actor, and further aims to maximize the entropy of this actor with an entropy maximization objective. We find that this actually results in a considerably more stable and scalable algorithm that, in practice, exceeds both the efficiency and final performance of DDPG. A similar method can be derived as a zero-step special case of stochastic value gradients (SVG(0))~\cite{heess2015learning}. However, SVG(0) differs from our method in that it optimizes the standard maximum expected return objective.

Maximum entropy reinforcement learning optimizes policies to maximize both the expected return and the expected entropy of the policy. This framework has been used in many contexts, from inverse reinforcement learning~\citep{ziebart2008maximum} to optimal control~\citep{todorov2008general,toussaint2009robot,rawlik2012stochastic}. Maximum a posteriori policy optimization (MPO) makes use of the probablistic view and optimizes the standard RL objective via expectation maximization \cite{abdolmaleki2018maximum}. In guided policy search \citep{levine2013guided,levine2016end}, the maximum entropy distribution is used to guide policy learning towards high-reward regions. More recently, several papers have noted the connection between Q-learning and policy gradient methods in the framework of maximum entropy learning~\citep{o2016pgq,haarnoja2017reinforcement,nachum2017bridging,schulman2017equivalence}. While most of the prior model-free works assume a discrete action space, \citet{nachum2017trust} approximate the maximum entropy distribution with a Gaussian, and \citet{haarnoja2017reinforcement} with a sampling network trained to draw samples from the optimal policy. Although the soft Q-learning algorithm proposed by \citet{haarnoja2017reinforcement} has a value function and actor network, it is not a true actor-critic algorithm: the Q-function is estimating the optimal Q-function, and the actor does not directly affect the Q-function except through the data distribution. Hence, \citet{haarnoja2017reinforcement} motivates the actor network as an approximate sampler, rather than the actor in an actor-critic algorithm. Crucially, the convergence of this method hinges on how well this sampler approximates the true posterior. In contrast, we prove that our method converges to the optimal policy from a given policy class, regardless of the policy parameterization. Furthermore, these prior maximum entropy methods generally do not exceed the performance of state-of-the-art off-policy algorithms, such as TD3 \cite{fujimoto2018addressing} or MPO \cite{abdolmaleki2018maximum}, when learning from scratch, though they may have other benefits, such as improved exploration and ease of fine-tuning.

\section{Preliminaries}
\label{sec:preliminaries}

We first introduce notation and summarize the standard and maximum entropy reinforcement learning frameworks.

\subsection{Notation}
We will address learning of maximum entropy policies in continuous action spaces. Our reinforcement learning problem can be defined as policy search in an a Markov decision process (MDP), defined by a tuple $(\sspace, \aspace, \pdyn, \reward)$. The state space $\sspace$ and action space $\aspace$ are assumed to be continuous, and the state transition probability $\pdyn:\ \sspace \times \sspace \times \aspace \rightarrow [0,\, \infty)$ represents the probability density of the next state $\stp\in\sspace$ given the current state $\st\in\sspace$ and action $\at\in\aspace$. The environment emits a reward $\reward: \sspace \times \aspace \rightarrow  [\rmin,\rmax]$ on each transition. We will also use $\rho_\policy(\st)$ and $\rho_\policy(\st,\at)$ to denote the state and state-action marginals of the trajectory distribution induced by a policy $\policy(\at|\st)$.

\subsection{Maximum Entropy Reinforcement Learning}
The standard reinforcement learning objective is the expected sum of rewards $\sum_t \E{(\st,\at)\sim\rho_\policy}{\reward(\st,\at)}$ and our goal is to learn a policy $\policy(\at|\st)$ that maximizes that objective. 
 The maximum entropy objective (see \eg \cite{ziebart2010modeling} generalizes the standard objective by augmenting it with an entropy term, such that the optimal policy additionally aims to maximize its entropy at each visited state:
\begin{align}
\label{eq:sql:maxent_objective}
\policy\opt = \arg\max_{\policy} \sum_{t} \E{(\st, \at) \sim \rho_\policy}{\reward(\st,\at) + \alpha\ent(\policy(\voidarg|\st))},
\end{align}
where $\alpha$ is the temperature parameter that determines the relative importance of the entropy term versus the reward, and thus controls the stochasticity of the optimal policy. Although the maximum entropy objective differs from the standard maximum expected return objective used in conventional reinforcement learning, the conventional objective can be recovered in the limit as $\alpha \rightarrow 0$. If we wish to extend either the conventional or the maximum entropy RL objective to infinite horizon problems, it is convenient to also introduce a discount factor $\discount$ to ensure that the sum of expected rewards (and entropies) is finite. In the context of policy search algorithms, the use of a discount factor is actually a somewhat nuanced choice, and writing down the precise objective that is optimized when using the discount factor is non-trivial \cite{thomas2014bias}. We include the discounted, infinite-horizon objective in \aref{app:infinite_horizon_objective}, but we will use the discount $\discount$ in the following derivations and in our final algorithm.

The maximum entropy objective has a number of conceptual and practical advantages. First, the policy is incentivized to explore more widely, while giving up on clearly unpromising avenues. Second, the policy can capture multiple modes of near-optimal behavior. In problem settings where multiple actions seem equally attractive, the policy will commit equal probability mass to those actions. In practice, we observe improved exploration with this objective, as also has been reported in the prior work \citep{schulman2017equivalence}, and we observe that it considerably improves learning speed over state-of-art methods that optimize the conventional RL objective function. 

Optimization problems of this type have been explored in a number of prior works~\cite{kappen2005path,todorov2007linearly,ziebart2008maximum}. These prior methods have proposed directly solving for the optimal Q-function, from which the optimal policy can be recovered~\citep{ziebart2008maximum,fox2015taming,haarnoja2017reinforcement}. In the following section, we discuss how we can devise a soft actor-critic algorithm through a policy iteration formulation, where we instead evaluate the Q-function of the current policy and update the policy through an \emph{off-policy} gradient update. Though such algorithms have previously been proposed for conventional reinforcement learning, our method is, to our knowledge, the first off-policy actor-critic method in the maximum entropy reinforcement learning framework.

\section{From Soft Policy Iteration to Soft Actor-Critic} 
\label{sec:soft_policy_iteration}
Our off-policy soft actor-critic algorithm can be derived starting from a maximum entropy variant of the policy iteration method. We will first present this derivation, verify that the corresponding algorithm converges to the optimal policy from its density class, and then present a practical deep reinforcement learning algorithm based on this theory. In this section, we treat the temperature as a constant, and later in \autoref{sec:entropy_constraint} propose an extension to SAC that adjusts the temperature automatically to match an entropy target in expectation.

\subsection{Soft Policy Iteration}
We will begin by deriving soft policy iteration, a general algorithm for learning optimal maximum entropy policies that alternates between policy evaluation and policy improvement in the maximum entropy framework. Our derivation is based on a tabular setting, to enable theoretical analysis and convergence guarantees, and we extend this method into the general continuous setting in the next section. We will show that soft policy iteration converges to the optimal policy within a set of policies which might correspond, for instance, to a set of parameterized densities.

In the policy evaluation step of soft policy iteration, we wish to compute the value of a policy $\policy$ according to the maximum entropy objective. For a fixed policy, the soft Q-value can be computed iteratively, starting from any function $Q: \sspace\times \aspace \rightarrow \reals$ and repeatedly applying a modified Bellman backup operator $\mathcal{T}^\policy$ given by
\begin{align}
\label{eq:soft_bellman_backup_op}
\mathcal{T}^\policy Q(\st, \at) \triangleq  \reward(\st, \at) + \discount \E{\stp \sim \pdyn}{V(\stp)},
\end{align}
where
\begin{align}
V(\st) = \E{\at\sim\policy}{\Q(\st, \at) - \alpha\log\policy(\at|\st)}
\label{eq:soft_value_function}
\end{align}
is the soft state value function. We can obtain the soft Q-function for any policy $\policy$ by repeatedly applying $\mathcal{T}^\policy$ as formalized below.
\begin{lemma}[Soft Policy Evaluation]
\label{lem:soft_policy_evaluation}
Consider the soft Bellman backup operator $\mathcal{T}^\policy$ in \autoref{eq:soft_bellman_backup_op} and a mapping $Q^0: \sspace \times \aspace\rightarrow \reals$ with $|\aspace|<\infty$, and define $\Q^{k+1} = \mathcal{T}^\policy \Q^k$. Then the sequence $Q^k$ will converge to the soft Q-function of $\policy$ as $k\rightarrow \infty$.
\begin{proof}
See \aref{app:lem_soft_policy_evaluation}.
\end{proof}
\end{lemma}
In the policy improvement step, we update the policy towards the exponential of the new soft Q-function. This particular choice of update can be guaranteed to result in an improved policy in terms of its soft value. Since in practice we prefer policies that are tractable, we will additionally restrict the policy to some set of policies $\Pi$, which can correspond, for example, to a parameterized family of distributions such as Gaussians.  To account for the constraint that $\policy \in \Pi$, we project the improved policy into the desired set of policies. While in principle we could choose any projection, it will turn out to be convenient to use the information projection defined in terms of the Kullback-Leibler divergence. In the other words, in the policy improvement step, for each state, we update the policy according to
\begin{align}
\policy_\mathrm{new} = \arg\underset{\policy'\in \Pi}{\min}\kl{\policy'(\voidarg|\st)}{\frac{\exp\left(\frac{1}{\alpha}Q^{\policy_\mathrm{old}}(\st, \voidarg)\right)}{Z^{\policy_\mathrm{old}}(\st)}}.
\label{eq:constrainted_policy_fitting}
\end{align}
The partition function $Z^{\policy_\mathrm{old}}(\st)$ normalizes the distribution, and while it is intractable in general, it does not contribute to the gradient with respect to the new policy and can thus be ignored. For this projection, we can show that the new, projected policy has a higher value than the old policy with respect to the maximum entropy objective. We formalize this result in \autoref{lem:policy_improvement}.
\begin{lemma}[Soft Policy Improvement]
\label{lem:policy_improvement}
Let $\policy_\mathrm{old} \in \Pi$ and let $\policy_\mathrm{new}$ be the optimizer of the minimization problem defined in \autoref{eq:constrainted_policy_fitting}. Then $\Q^{\policy_\mathrm{new}}(\st, \at) \geq \Q^{\policy_\mathrm{old}}(\st, \at)$ for all $(\st, \at) \in \sspace\times\aspace$ with $|\aspace|<\infty$.
\begin{proof}
See \aref{app:lem_policy_improvement}.
\end{proof}
\end{lemma}
The full soft policy iteration algorithm alternates between the soft policy evaluation and the soft policy improvement steps, and it will provably converge to the optimal maximum entropy policy among the policies in $\Pi$ (\autoref{the:soft_policy_iteration}). Although this algorithm will provably find the optimal solution, we can perform it in its exact form only in the tabular case. Therefore, we will next approximate the algorithm for continuous domains, where we need to rely on a function approximator to represent the Q-values, and running the two steps until convergence would be computationally too expensive. The approximation gives rise to a new practical algorithm, called soft actor-critic.
\begin{theorem}[Soft Policy Iteration]
\label{the:soft_policy_iteration}
Repeated application of soft policy evaluation and soft policy improvement from any $\policy\in\Pi$ converges to a policy $\policy\opt$ such that $Q^{\policy\opt}(\st, \at) \geq Q^{\policy}(\st, \at)$ for all $\policy \in \Pi$ and $(\st, \at) \in \sspace\times\aspace$, assuming $|\aspace|<\infty$.
\begin{proof}
See \aref{app:the_soft_policy_iteration}.
\end{proof}
\end{theorem}

\subsection{Soft Actor-Critic}
As discussed above, large continuous domains require us to derive a practical approximation to soft policy iteration. To that end, we will use function approximators for both the soft Q-function and the policy, and instead of running evaluation and improvement to convergence, alternate between optimizing both networks with stochastic gradient descent.
We will consider a parameterized soft Q-function $\Q_\params(\st, \at)$ and a tractable policy $\policy_\pparams(\at|\st)$. The parameters of these networks are  $\params$ and $\pparams$. For example, the soft Q-function can be modeled as expressive neural networks, and the policy as a Gaussian with mean and covariance given by neural networks. We will next derive update rules for these parameter vectors.

The soft Q-function parameters can be trained to minimize the soft Bellman residual
\begin{align}
J_\Q(\params) = \E{(\st, \at)\sim\mathcal{D}}{\frac{1}{2}\left(\Q_\params(\st, \at) - \left(\reward(\st, \at) + \discount \E{\stp\sim\pdyn}{V_{\bar\params}(\stp)}\right)\right)^2},
\label{eq:q_cost}
\end{align}
where the value function is implicitly parameterized through the soft Q-function parameters via \autoref{eq:soft_value_function}, and it can be optimized with stochastic gradients\footnote{In \cite{haarnoja2018soft} we introduced an additional function approximator for the value function, but later found it to be unnecessary.}
\begin{align}
\hat \nabla_\params J_Q(\params) =  \nabla_\params \Q_\params(\at, \st) \left(\Q_\params(\st, \at) - \left(r(\st,\at) + \gamma \left(Q_{\bar\params}(\stp, \atp) - \alpha \log\left(\pi_\pparams(\atp|\stp\right)\right)\right)\right).
\label{eq:q_stochastic_gradient}
\end{align}
The update makes use of a target soft Q-function with parameters $\bar\params$ that are obtained as an exponentially moving average of the soft Q-function weights, which has been shown to stabilize training~\citep{mnih2015human}. Finally, the policy parameters can be learned by directly minimizing the expected KL-divergence in \autoref{eq:constrainted_policy_fitting} (multiplied by $\alpha$ and ignoring the constant log-partition function and by $\alpha$):
\begin{align}
J_\policy(\pparams) = \E{\st\sim\mathcal{D}}{\E{\at\sim\policy_\pparams}{\alpha \log\left(\policy_\pparams(\at|\st)\right) - Q_\params(\st, \st)}}
\label{eq:policy_objective}
\end{align}
There are several options for minimizing $J_\policy$. A typical solution for policy gradient methods is to use the likelihood ratio gradient estimator~\citep{williams1992simple}, which does not require backpropagating the gradient through the policy and the target density networks. However, in our case, the target density is the Q-function, which is represented by a neural network an can be differentiated, and it is thus convenient to apply the reparameterization trick instead, resulting in a lower variance estimator. To that end, we reparameterize the policy using a neural network transformation 
\begin{align}
\at = f_\pparams(\epsilon_t; \st),
\end{align}
where $\epsilon_t$ is an input noise vector, sampled from some fixed distribution, such as a spherical Gaussian. We can now rewrite the objective in~\autoref{eq:policy_objective} as
\begin{align}
J_\policy(\pparams) = \E{\st\sim\mathcal{D},\epsilon_t\sim\gauss}{\alpha \log \policy_\pparams(f_\pparams(\epsilon_t;\st)|\st) - Q_\params(\st, f_\pparams(\epsilon_t;\st))},
\label{eq:reparam_objective}
\end{align}
where $\policy_\pparams$ is defined implicitly in terms of $f_\pparams$. We can approximate the gradient of~\autoref{eq:reparam_objective} with
\begin{align}
&\hat\nabla_\pparams J_\policy(\pparams) = \nabla_\pparams \alpha \log\left( \policy_\pparams(\at|\st)\right) + (\nabla_\at \alpha \log \left(\policy_\pparams(\at|\st)\right)
- \nabla_\at Q(\st, \at))\nabla_\pparams f_\pparams(\epsilon_t;\st),
\label{eq:policy_gradient}
\end{align}
where $\at$ is evaluated at $f_\pparams(\epsilon_t; \st)$. This unbiased gradient estimator extends the DDPG style policy gradients~\citep{lillicrap2015continuous} to any tractable stochastic policy.

\section{Automating Entropy Adjustment for Maximum Entropy RL}
\label{sec:entropy_constraint}
In the previous section, we derived a practical off-policy algorithm for learning maximum entropy policies of a given temperature. Unfortunately, choosing the optimal temperature is non-trivial, and the temperature needs to be tuned for each task. Instead of requiring the user to set the temperature manually, we can automate this process by formulating a different maximum entropy reinforcement learning objective, where the entropy is treated as a constraint. The magnitude of the reward differs not only across tasks, but it also depends on the policy, which improves over time during training. Since the optimal entropy depends on this magnitude, this makes the temperature adjustment particularly difficult: the entropy can vary unpredictably both across tasks and during training as the policy becomes better. Simply forcing the entropy to a fixed value is a poor solution, since the policy should be free to explore more in regions where the optimal action is uncertain, but remain more deterministic in states with a clear distinction between good and bad actions. Instead, we formulate a constrained optimization problem where the average entropy of the policy is constrained, while the entropy at different states can vary. Similar approach was taken in \cite{abdolmaleki2018maximum}, where the policy was constrained to remain close to the previous policy. We show that the dual to this constrained optimization leads to the soft actor-critic updates, along with an additional update for the dual variable, which plays the role of the temperature. Our formulation also makes it possible to learn the entropy with more expressive policies that can model multi-modal distributions, such as policies based on normalizing flows \cite{haarnoja2018latent} for which no closed form expression for the entropy exists. We will derive the update for finite horizon case, and then derive an approximation for stationary policies by dropping the time dependencies from the policy, soft Q-function, and the temperature.

Our aim is to find a stochastic policy with maximal expected return that satisfies a minimum expected entropy constraint. Formally, we want to solve the constrained optimization problem
\begin{align}
    \max_{\pi_{0:T}} \E{\rho_\pi}{\sum_{t=0}^T r(\st,\at)} \text{ s.t. } \E{(\st, \at)\sim\rho_\pi}{-\log(\pi_t(\at|\st))} \geq \entropy\ \ \forall t
    \label{eq:ecsac:contrained_rl}
\end{align}
where $\ent$ is a desired minimum expected entropy. Note that, for fully observed MDPs, the policy that optimizes the expected return is deterministic, so we expect this constraint to usually be tight and do not need to impose an upper bound on the entropy.

Since the policy at time $t$ can only affect the future objective value, we can employ an (approximate) dynamic programming approach, solving for the policy backward through time. We rewrite the objective as an iterated maximization
\begin{equation}
    \max_{\pi_0} \left( \E{}{r(\state_0, \action_0)} + \max_{\pi_1} \left( \E{}{ \ldots } + \max_{\pi_T} \E{}{r(\state_T, \action_T)} \right) \right),
\end{equation}
subject to the constraint on entropy. Starting from the last time step, we change the constrained maximization to the dual problem. Subject to $\E{(\sT, \aT)\sim\rho_\pi}{-\log(\pi_T(\sT|\sT))} \geq \entropy$,
\begin{align}
    \max_{\pi_T} &\E{(\st, \at)\sim\rho_\pi}{r(\state_T, \action_T)} = \min_{\alpha_T \geq 0} \max_{\pi_T} \E{}{r(\state_T, \action_T) - \alpha_T \log \policy(\action_T|\state_T)} -\alpha_T \entropy,
\end{align}
where $\alpha_T$ is the dual variable. We have also used strong duality, which holds since the objective is linear and the constraint (entropy) is convex function in $\policy_T$. This dual objective is closely related to the maximum entropy objective with respect to the policy, and the optimal policy is the maximum entropy policy corresponding to temperature $\alpha_T$: $\pi_T^*(\action_T | \state_T; \alpha_T)$. We can solve for the optimal dual variable $\alpha_T\opt$ as
\begin{align}
    \arg \min_{\alpha_T} \E{\st, \at\sim \policy_t\opt}{- \alpha_T\log\pi_T\opt(\action_T|\state_T; \alpha_T) - \alpha_T \ent}.
\end{align}
To simplify notation, we make use of the recursive definition of the soft Q-function
\begin{align}
   Q_t\opt(\st, \at; \pi\opt_{t+1:T}, \alpha^*_{t+1:T}) = \E{}{r(\st, \at)} + \E{\rho_\pi}{Q_{t+1}\opt (\state_{t+1}, \action_{t+1}) - \alpha_{t+1}\opt\log \pi_{t+1}\opt(\action_{t+1}|\state_{t+1})},
\end{align}
with $Q\opt_{T}(\state_T, \action_T) = \E{}{r(\state_T, \action_T)}$. Now, subject to the entropy constraints and again using the dual problem, we have
\begin{align}
    \max_{\pi_{T-1}} &\left( \E{}{r(\state_{T-1}, \action_{T-1}) } + \max_{\pi_T} \E{}{r(\state_T, \action_T)} \right)\\
    &= \max_{\pi_{T-1}} \left( Q\opt_{T-1}(\state_{T-1}, \action_{T-1}) - \alpha_T \entropy \right)\notag\\
    &= \min_{\alpha_{T-1} \geq 0} \max_{\pi_{T-1}} \bigg( \E{}{Q\opt_{T-1}(\state_{T-1}, \action_{T-1})} - \E{}{\alpha_{T-1} \log \policy(\action_{T-1}|\state_{T-1})} - \alpha_{T-1}  \entropy \bigg) + \alpha\opt_T \entropy\notag.
\end{align}
In this way, we can proceed backwards in time and recursively optimize \autoref{eq:ecsac:contrained_rl}. Note that the optimal policy at time $t$ is a function of the dual variable $\alpha_t$. Similarly, we can solve the optimal dual variable $\alpha_t\opt$ after solving for $Q\opt_t$ and $\policy_t\opt$:
\begin{align}
\alpha_t\opt = \arg \min_{\alpha_t} \E{\at\sim \policy_t\opt}{- \alpha_t\log\pi_t\opt(\at|\st; \alpha_t) - \alpha_t \bar\ent}.
\label{eq:optimal_temperature}
\end{align}
The solution in \autoref{eq:optimal_temperature} along with the policy and soft Q-function updates described in \autoref{sec:soft_policy_iteration} constitute the core of our algorithm, and in theory, exactly solving them recursively optimize the optimal entropy-constrained maximum expected return objective in~\autoref{eq:ecsac:contrained_rl}, but in practice, we will need to resort to function approximators and stochastic gradient descent.

\section{Practical Algorithm}     
Our algorithm makes use of two soft Q-functions to mitigate  positive bias in the policy improvement step that is known to degrade performance of value based methods~\citep{hasselt2010double,fujimoto2018addressing}. In particular, we parameterize two soft Q-functions, with parameters $\params_i$, and train them independently to optimize $J_Q(\params_i)$. We then use the minimum of the the soft Q-functions for the stochastic gradient in \autoref{eq:q_stochastic_gradient} and policy gradient in \autoref{eq:policy_gradient}, as proposed by \citet{fujimoto2018addressing}. Although our algorithm can learn challenging tasks, including a 21-dimensional Humanoid, using just a single Q-function, we found two soft Q-functions significantly speed up training, especially on harder tasks. 

In addition to the soft Q-function and the policy, we also learn  $\alpha$
by minimizing the dual objective in~\autoref{eq:optimal_temperature}. This can be done by approximating dual gradient descent~\cite{boyd2004convex}. Dual gradient descent alternates between optimizing the Lagrangian with respect to the primal variables to convergence, and then taking a gradient step on the dual variables. While optimizing with respect to the primal variables fully is impractical, a truncated version that performs incomplete optimization (even for a single gradient step) can be shown to converge under convexity assumptions~\cite{boyd2004convex}. While such assumptions do not apply to the case of nonlinear function approximators such as neural networks, we found this approach to still work in practice. Thus, we compute gradients for $\alpha$ with the following objective:
\begin{align}
J(\alpha)  = \E{\at\sim \policy_t}{ - \alpha\log\pi_t(\at|\st) - \alpha \bar\ent}.
\label{eq:ecsac:alpha_objective}
\end{align}

The final algorithm is listed in \autoref{alg:soft_actor_critic}. The method alternates between collecting experience from the environment with the current policy and updating the function approximators using the stochastic gradients from batches sampled from a replay pool. Using off-policy data from a replay pool is feasible because both value estimators and the policy can be trained entirely on off-policy data. The algorithm is agnostic to the parameterization of the policy, as long as it can be evaluated for any arbitrary state-action tuple.

\begin{algorithm}[tb]
\caption{Soft Actor-Critic}
\label{alg:soft_actor_critic}
\begin{algorithmic}
\Require $\params_1$, $\params_2$, $\pparams$ \Comment{Initial parameters}
\State $\bar \params_1 \leftarrow \params_1$, $\bar \params_2 \leftarrow \params_2$ \Comment{Initialize target network weights}
\State $\mathcal{D}\leftarrow\emptyset$ \Comment{Initialize an empty replay pool}
\For{each iteration}
	\For{each environment step}
	    \State $\at \sim \policy_\pparams(\at|\st)$ \Comment{Sample action from the policy}
	    \State $\stp \sim \pdyn(\stp| \st, \at)$ \Comment{Sample transition from the environment}
	    \State $\mathcal{D} \leftarrow \mathcal{D} \cup \left\{(\st, \at, \reward(\st, \at), \stp)\right\}$ \Comment{Store the transition in the replay pool}
	\EndFor
	\For{each gradient step}
	    \State $\params_i \leftarrow \params_i - \lambda_Q \hat \nabla_{\params_i} J_\Q(\params_i)$ for $i\in\{1, 2\}$ \Comment{Update the Q-function parameters}
	    \State $\pparams \leftarrow \pparams - \lambda_\policy \hat \nabla_\pparams J_\policy(\pparams)$\Comment{Update policy weights}
	    \State $\alpha \leftarrow \alpha - \lambda \hat \nabla_\alpha J(\alpha)$ \Comment{Adjust temperature}
	    \State $\bar\params_i\leftarrow \tau \params_i + (1-\tau)\bar\params_i$ for $i\in\{1,2\}$\Comment{Update target network weights}
	\EndFor
\EndFor
\Ensure $\params_1$, $\params_2$, $\pparams$\Comment{Optimized parameters}
\end{algorithmic}
\end{algorithm}

\section{Experiments}
\label{sec:experiments}
The goal of our experimental evaluation is to understand how the sample complexity and stability of our method compares with prior off-policy and on-policy deep reinforcement learning algorithms. We compare our method to prior techniques on a range of challenging continuous control tasks from the OpenAI gym benchmark suite~\citep{brockman2016openai} and also on the rllab implementation of the Humanoid task~\citep{duan2016benchmarking}. Although the easier tasks can be solved by a wide range of different algorithms, the more complex benchmarks, such as the 21-dimensional Humanoid (rllab), are exceptionally difficult to solve with off-policy algorithms~\citep{duan2016benchmarking}. The stability of the algorithm also plays a large role in performance: easier tasks make it more practical to tune hyperparameters to achieve good results, while the already narrow basins of effective hyperparameters become prohibitively small for the more sensitive algorithms on the hardest benchmarks, leading to poor performance~\citep{gu2016q}.

\subsection{Simulated Benchmarks}
\begin{figure}[tb]
    \centering
    \begin{subfigure}{0.32\textwidth}
        \includegraphics[width=\textwidth, trim={0 0 5mm 0}, clip]{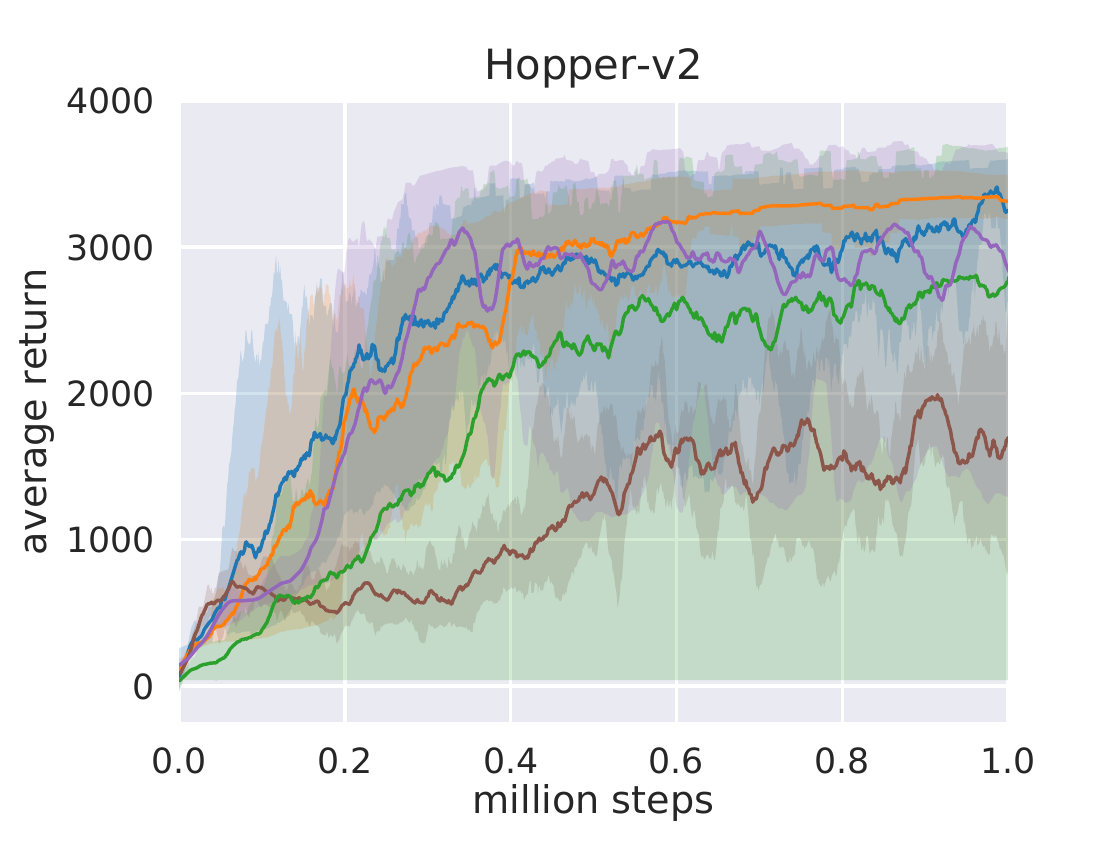}
    \end{subfigure}
    \begin{subfigure}{0.32\textwidth}
        \includegraphics[width=\textwidth, trim={0 0 5mm 0}, clip]{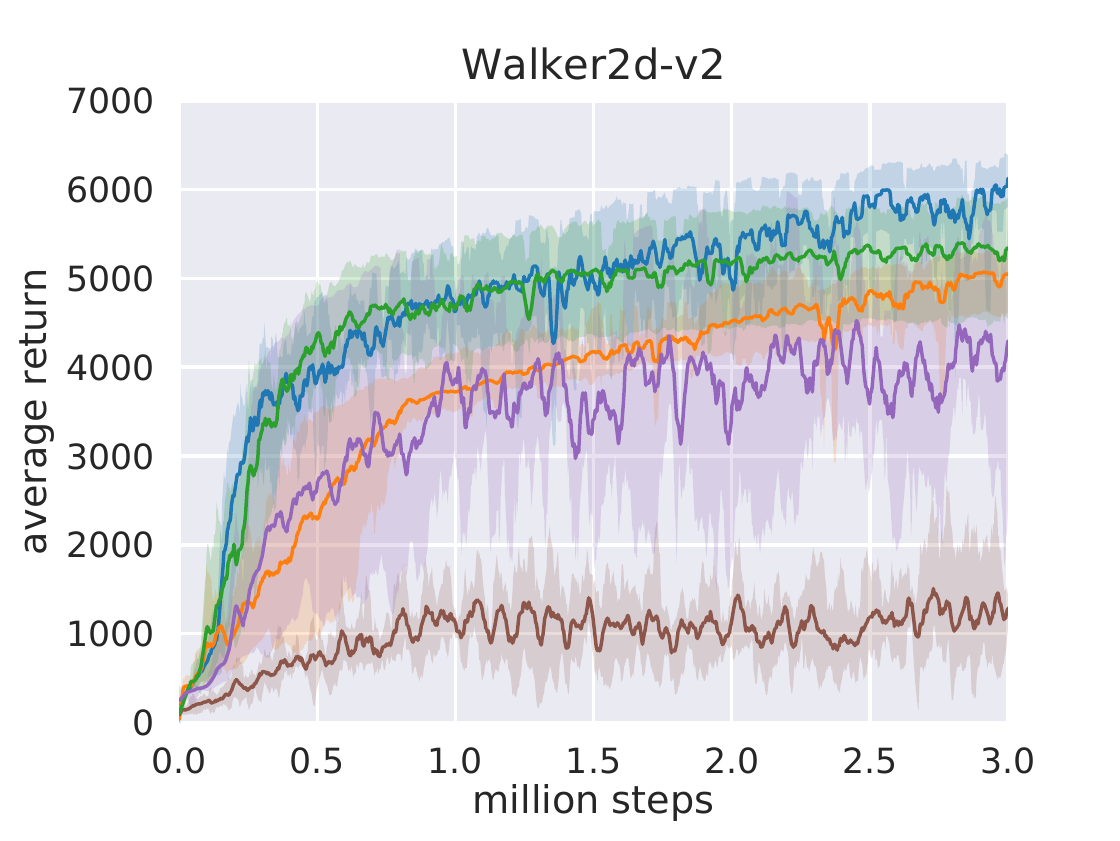}
    \end{subfigure}
    \begin{subfigure}{0.32\textwidth}
        \includegraphics[width=\textwidth, trim={0 0 5mm 0}, clip]{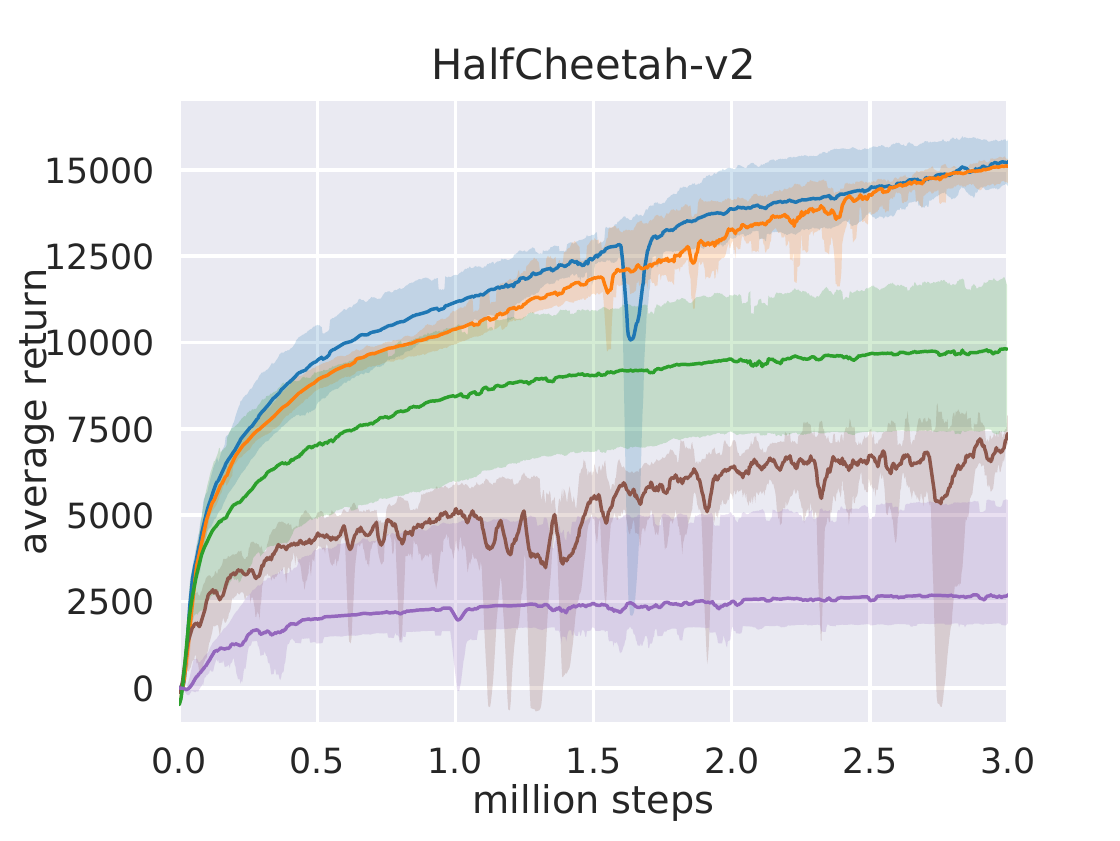}
    \end{subfigure}
    \begin{subfigure}{0.32\textwidth}
        \includegraphics[width=\textwidth, trim={0 0 5mm 0}, clip]{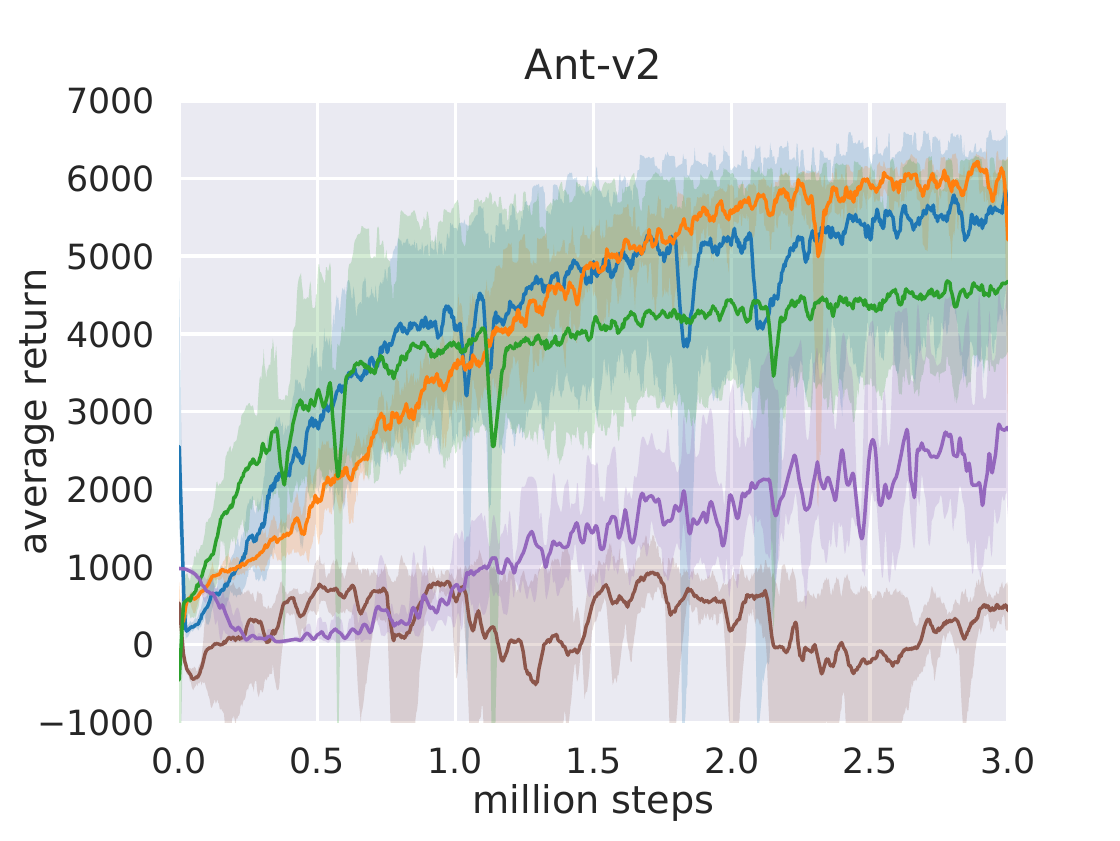}
    \end{subfigure}
    \begin{subfigure}{0.32\textwidth}
        \includegraphics[width=\textwidth, trim={0 0 5mm 0}, clip]{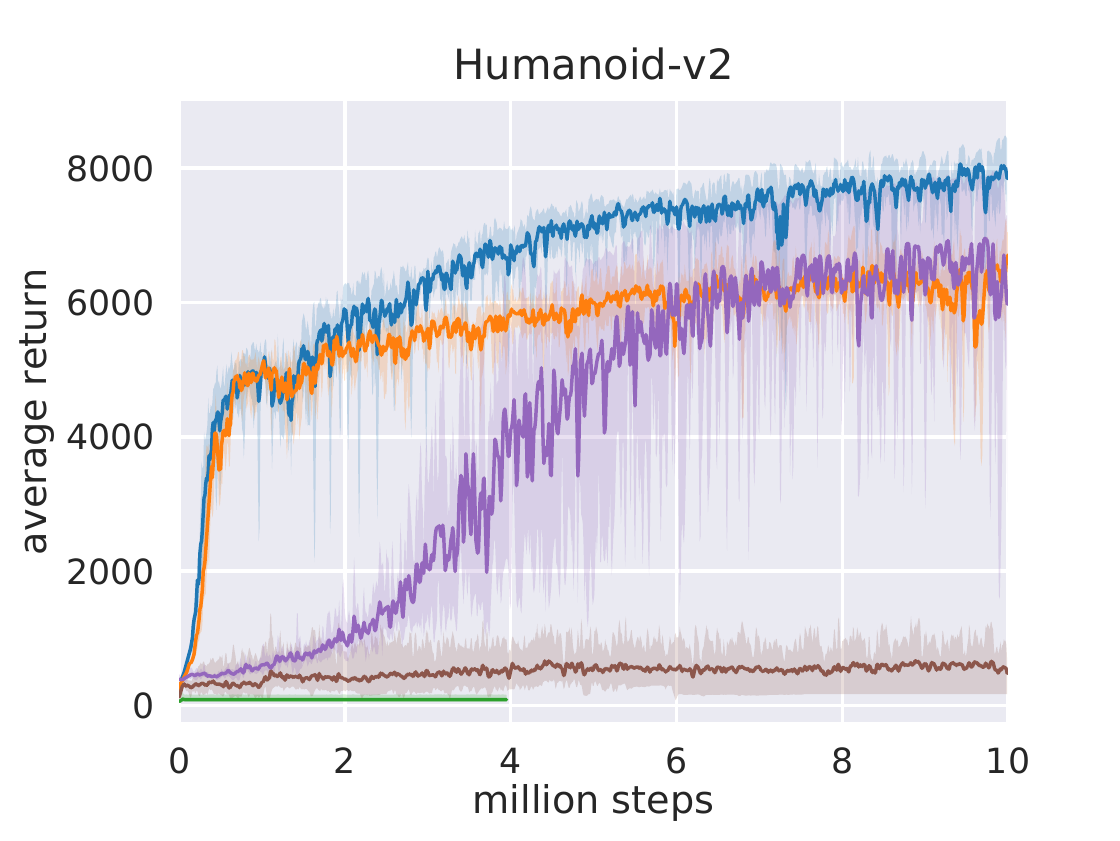}
    \end{subfigure}
    \begin{subfigure}{0.32\textwidth}
        \includegraphics[width=\textwidth, trim={0 0 5mm 0}, clip]{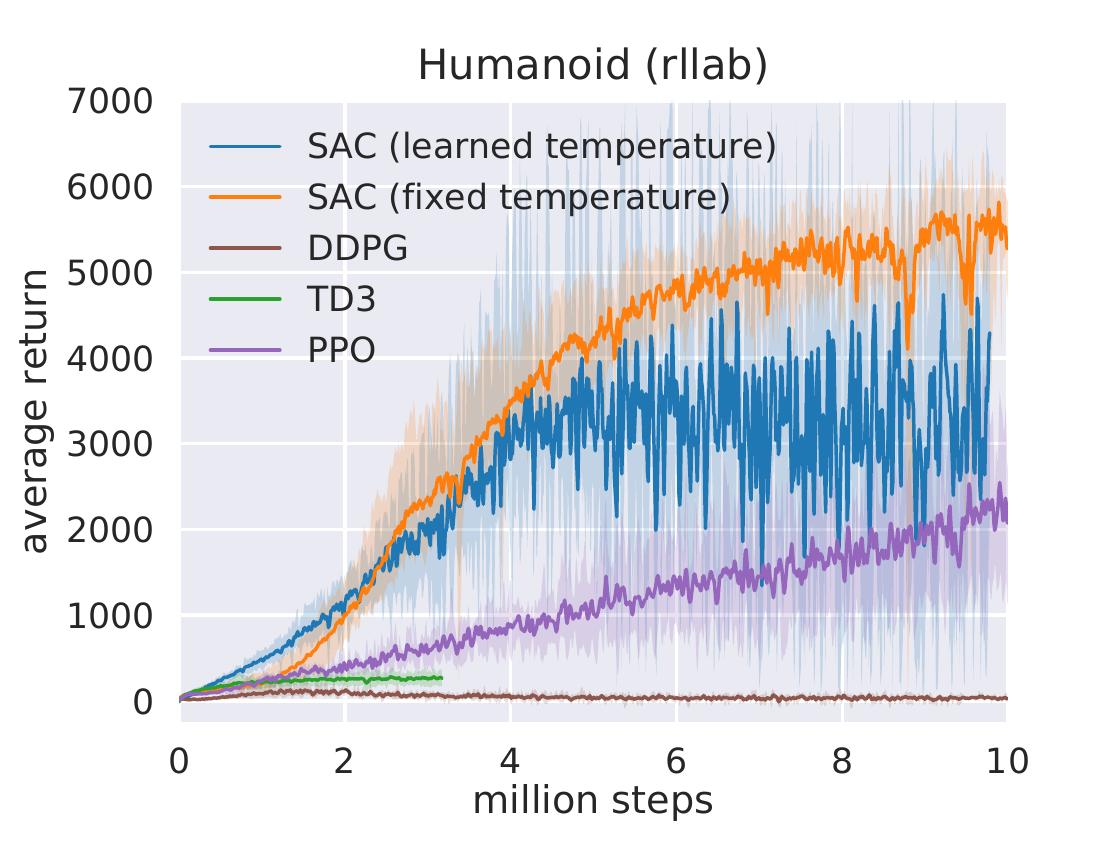}
    \end{subfigure}
    \caption{\small Training curves on continuous control benchmarks. Soft actor-critic (blue and yellow) performs consistently across all tasks and outperforming both on-policy and off-policy methods in the most challenging  tasks.}
	\label{fig:training_curves}
\end{figure}

We compare our method to deep deterministic policy gradient (DDPG)~\citep{lillicrap2015continuous}, an algorithm that is regarded as one of the more efficient off-policy deep RL methods~\citep{duan2016benchmarking}; proximal policy optimization (PPO)~\citep{schulman2017proximal}, a stable and effective on-policy policy gradient algorithm; and soft Q-learning (SQL)~\citep{haarnoja2017reinforcement}, a recent off-policy algorithm for learning maximum entropy policies. Our SQL implementation also includes two Q-functions, which we found to improve its performance in most environments.  We additionally compare to twin delayed deep deterministic policy gradient algorithm (TD3) \citep{fujimoto2018addressing}, using the author-provided implementation. This is an extension to DDPG, proposed concurrently to our method, that first applied the double Q-learning trick to continuous control along with other improvements. We turned off the exploration noise for evaluation for DDPG and PPO. For maximum entropy algorithms, which do not explicitly inject exploration noise, we either evaluated with the exploration noise (SQL) or use the mean action (SAC). The source code of our SAC implementation\footnote{\href{http://github.com/rail-berkeley/softlearning/}{http://github.com/rail-berkeley/softlearning/}} is available online. %

\autoref{fig:training_curves} shows the total average return of evaluation rollouts during training for DDPG, PPO, and TD3. We train five different instances of each algorithm with different random seeds, with each performing one evaluation rollout every 1000 environment steps. The solid curves corresponds to the mean and the shaded region to the minimum and maximum returns over the five trials. For SAC, we include both a version, where the temperature parameter is fixed and treated as a hyperparameter and tuned for each environment separately (orange), and a version where the temperature is adjusted automatically (blue). The results show that, overall, SAC performs comparably to the baseline methods on the easier tasks and outperforms them on the harder tasks with a large margin, both in terms of learning speed and the final performance. For example, DDPG fails to make any progress on Ant-v1, Humanoid-v1, and Humanoid (rllab), a result that is corroborated by prior work~\citep{gu2016q,duan2016benchmarking}. SAC also learns considerably faster than PPO as a consequence of the large batch sizes PPO needs to learn stably on more high-dimensional and complex tasks. Another maximum entropy RL algorithm, SQL, can also learn all tasks, but it is slower than SAC and has worse asymptotic performance. The quantitative results attained by SAC in our experiments also compare very favorably to results reported by other methods in prior work~\citep{duan2016benchmarking,gu2016q,henderson2017deep}, indicating that both the sample efficiency and final performance of SAC on these benchmark tasks exceeds the state of the art. The results also indicate that the automatic temperature tuning scheme works well across all the environments, and thus effectively eliminates the need for tuning the temperature. All hyperparameters used in this experiment for SAC are listed in \aref{app:hypers}.

\subsection{Quadrupedal Locomotion in the Real World}
\label{sec:minitaur}

In this section, we describe an application of our method to learn walking gaits directly in the real world. We use the Minitaur robot, a small-scale quadruped with eight direct-drive actuators \cite{kenneally2016design}. Each leg is controlled by two actuators that allow it to move in the sagittal plane. The Minitaur is equipped with motor encoders that measure the motor angles and an IMU that measures orientation and angular velocity of Minitaur's base. The action space are the swing angle and the extension of each leg, which are then mapped to desired motor positions and tracked with  a PD controller \cite{tan2018sim}. The observations include the motor angles as well as roll and pitch angles and angular velocities of the base. We exclude yaw since it is unreliable due to drift and irrelevant for the walking task. Note that latencies and contacts in the system make the dynamics non-Markovian, which can significantly degrade learning performance. We therefore construct the state out of the current and past five observations and actions. The reward function  rewards large forward velocity, which is estimated using a motion capture system, and penalizes large angular accelerations, computed via finite differentiation from the last three actions. We also found it necessary to penalize for large pitch angles and for extending the front legs under the robot, which we found to be the most common failure cases that would require manual reset. We parameterize the policy and the value functions with feed-forward neural networks with two hidden-layers and 256 neurons per layer.

We have developed a semi-automatic robot training pipeline 
that consists of two components parallel jobs: training and data collection. These jobs run asynchronously on two different computers. The training process runs on a workstation, which updates the neural networks and periodically downloads the latest data from the robot and uploads the latest policy to the robot. On the robot, the on-board Nvidia Jetson TX2 runs the data collection job, which executes the policy, collects the trajectory and uploads these data to the workstation through Ethernet. Once the training is started, minimal human intervention is needed, except for the need to reset the robot state if it falls or drifts far from the initial state.

This learning task presents substantially challenges for real-world reinforcement learning. The robot is underactuated, and must therefore delicately balance contact forces on the legs to make forward progress. An untrained policy can lose balance and fall, and too many falls will eventually damage the robot, making sample-efficient learning essentially. Our method successfully learns to walk from 160k environment steps, or approximately 400 episodes with maximum length of 500 steps, which amount to approximately 2 hours of real-world training time.

\begin{figure}[tb]
    \centering
    
    \begin{subfigure}{0.195\textwidth}
        \includegraphics[width=\textwidth, trim={0 0 0 0}, clip]{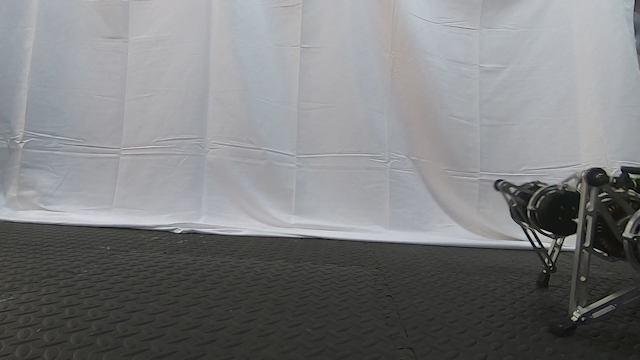}
    \end{subfigure}
    \hfill
    \begin{subfigure}{0.195\textwidth}
        \includegraphics[width=\textwidth, trim={0 0 0 0}, clip]{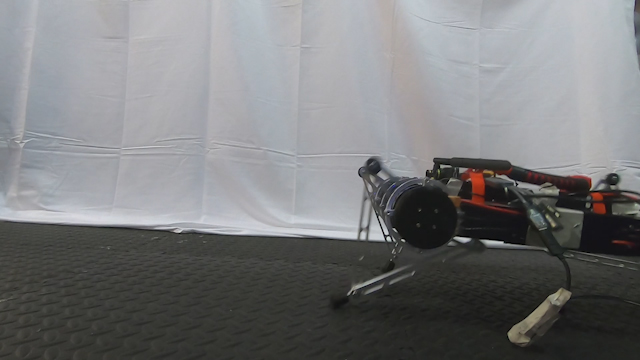}
    \end{subfigure}
    \hfill
    \begin{subfigure}{0.195\textwidth}
        \includegraphics[width=\textwidth, trim={0 0 0 0}, clip]{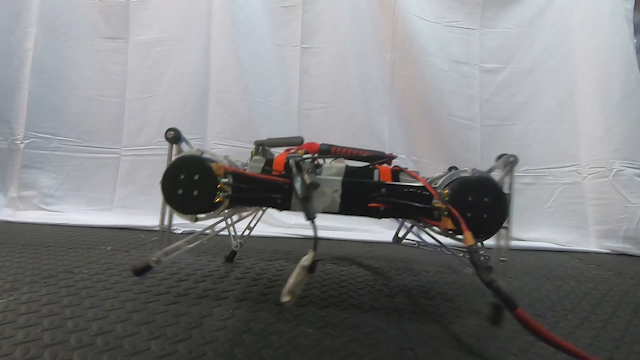}
    \end{subfigure}
    \hfill
    \begin{subfigure}{0.195\textwidth}
        \includegraphics[width=\textwidth, trim={0 0 0 0}, clip]{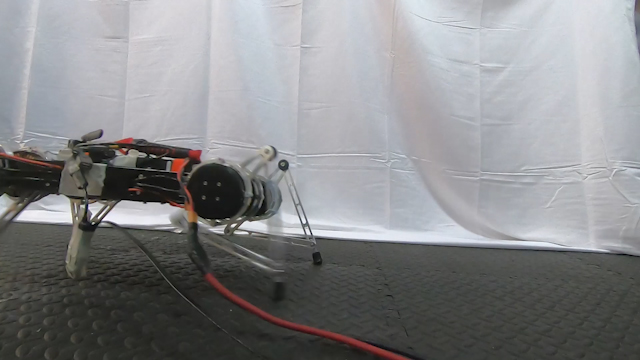}
    \end{subfigure}
    \hfill
    \begin{subfigure}{0.195\textwidth}
        \includegraphics[width=\textwidth, trim={0 0 0 0}, clip]{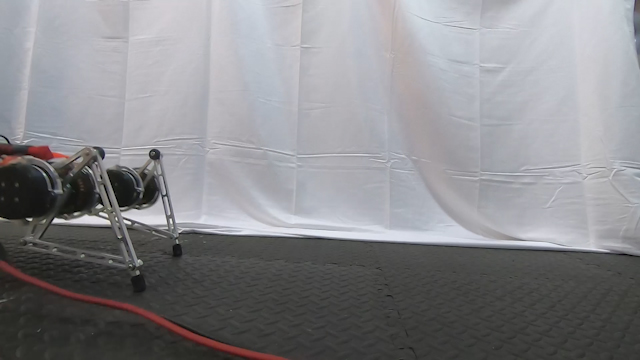}
    \end{subfigure}
    \hfill

    \begin{subfigure}{0.195\textwidth}
        \includegraphics[width=\textwidth, trim={0 0 0 0}, clip]{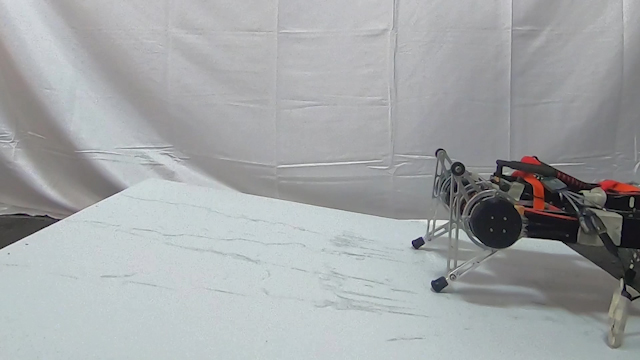}
    \end{subfigure}
    \hfill
    \begin{subfigure}{0.195\textwidth}
        \includegraphics[width=\textwidth, trim={0 0 0 0}, clip]{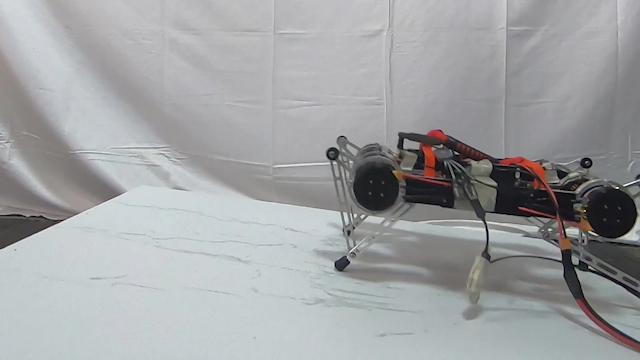}
    \end{subfigure}
    \hfill
    \begin{subfigure}{0.195\textwidth}
        \includegraphics[width=\textwidth, trim={0 0 0 0}, clip]{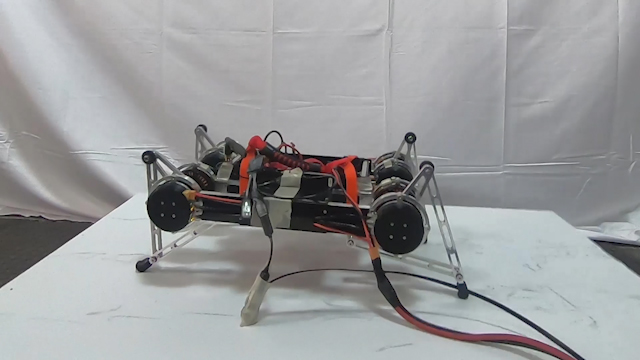}
    \end{subfigure}
    \hfill
    \begin{subfigure}{0.195\textwidth}
        \includegraphics[width=\textwidth, trim={0 0 0 0}, clip]{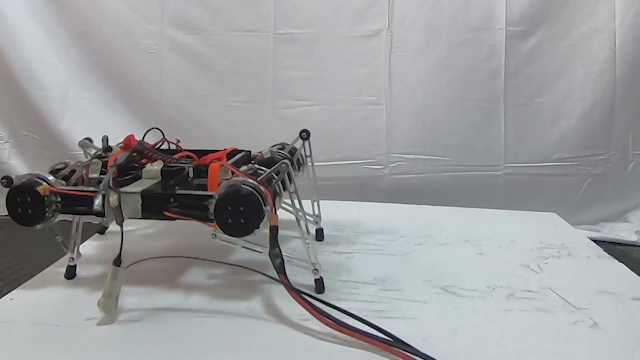}
    \end{subfigure}
    \hfill
    \begin{subfigure}{0.195\textwidth}
        \includegraphics[width=\textwidth, trim={0 0 0 0}, clip]{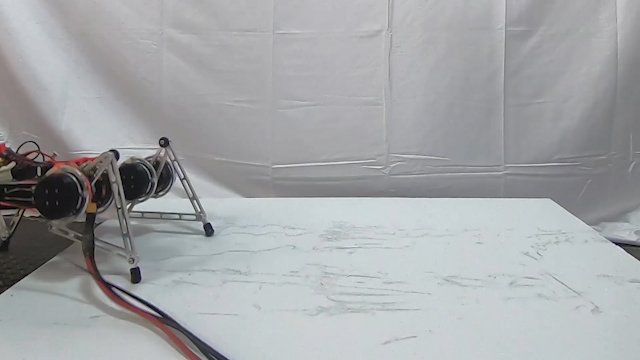}
    \end{subfigure}
    
    \begin{subfigure}{0.195\textwidth}
        \includegraphics[width=\textwidth, trim={0 0 0 0}, clip]{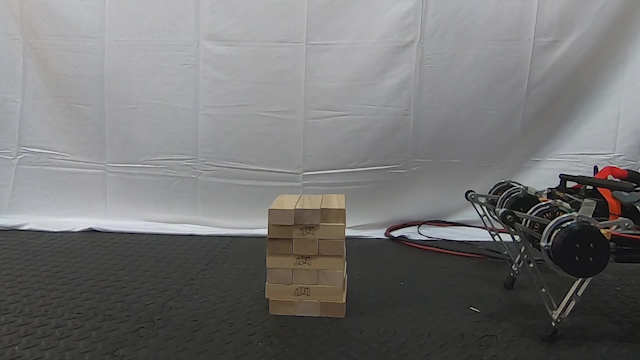}
    \end{subfigure}
    \hfill
    \begin{subfigure}{0.195\textwidth}
        \includegraphics[width=\textwidth, trim={0 0 0 0}, clip]{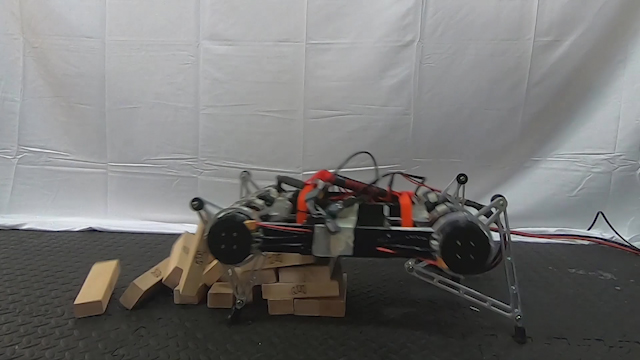}
    \end{subfigure}
    \hfill
    \begin{subfigure}{0.195\textwidth}
        \includegraphics[width=\textwidth, trim={0 0 0 0}, clip]{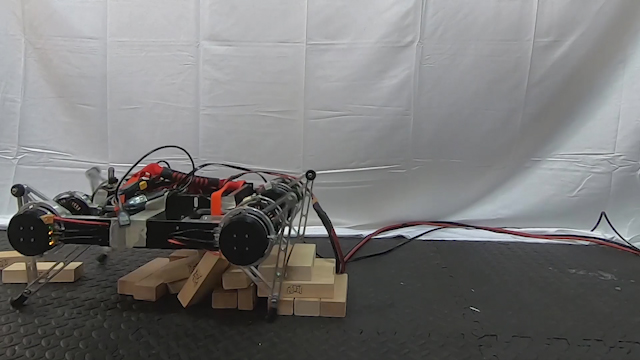}
    \end{subfigure}
    \hfill
    \begin{subfigure}{0.195\textwidth}
        \includegraphics[width=\textwidth, trim={0 0 0 0}, clip]{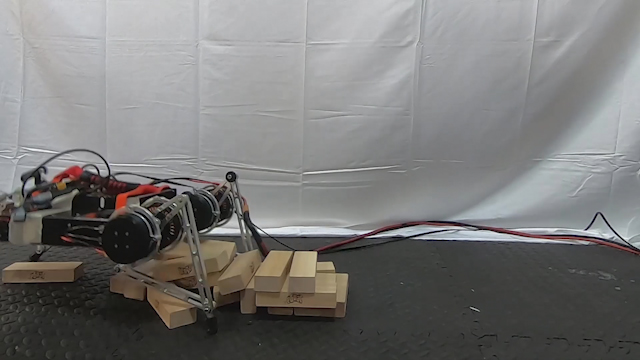}
    \end{subfigure}
    \hfill
    \begin{subfigure}{0.195\textwidth}
        \includegraphics[width=\textwidth, trim={0 0 0 0}, clip]{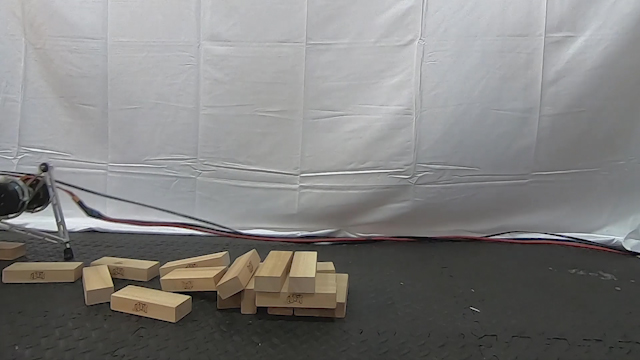}
    \end{subfigure}
    
    \begin{subfigure}{0.195\textwidth}
        \includegraphics[width=\textwidth, trim={0 0 0 0}, clip]{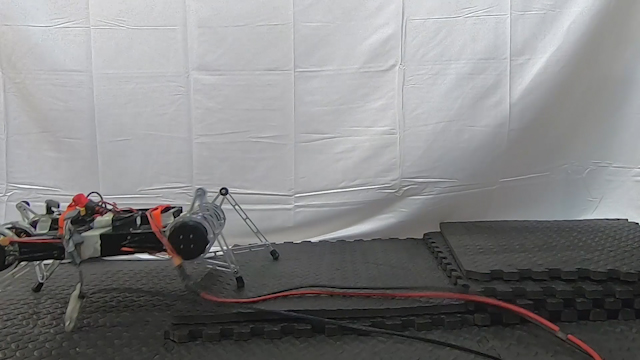}
    \end{subfigure}
    \hfill 
    \begin{subfigure}{0.195\textwidth}
        \includegraphics[width=\textwidth, trim={0 0 0 0}, clip]{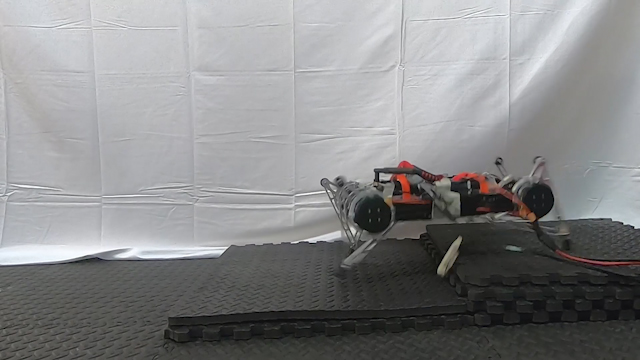}
    \end{subfigure}
    \hfill 
    \begin{subfigure}{0.195\textwidth}
        \includegraphics[width=\textwidth, trim={0 0 0 0}, clip]{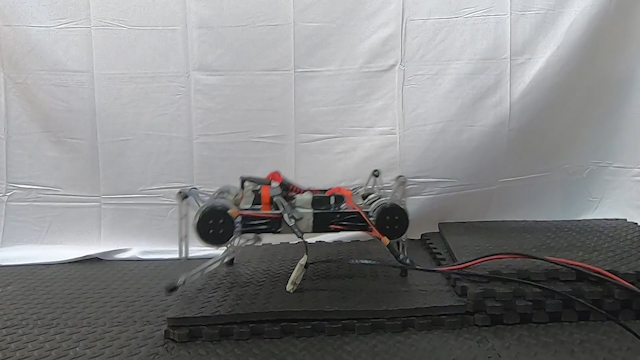}
    \end{subfigure}
    \hfill 
    \begin{subfigure}{0.195\textwidth}
        \includegraphics[width=\textwidth, trim={0 0 0 0}, clip]{figures/minitaur/steps/20181209_demo_step_2_clipped24.jpg}
    \end{subfigure}
    \hfill 
    \begin{subfigure}{0.195\textwidth}
        \includegraphics[width=\textwidth, trim={0 0 0 0}, clip]{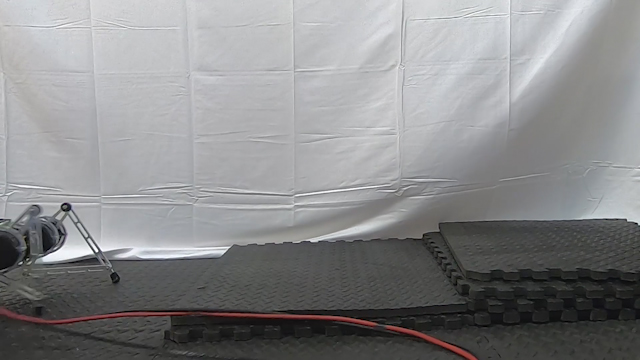}
    \end{subfigure}
    
    \caption{\small We trained the Minitaur robot to walk on flat terrain (first row) and tested how the learned gait generalizes to unseen situations (other rows)}
	\label{fig:minitaur_walking}	
\end{figure}

However, in the real world, the utility of a locomotion policy hinges critically on its ability to generalize to different terrains and obstacles. Although we trained our policy only on flat terrain, as illustrated in \autoref{fig:minitaur_walking} (first row), we then tested it on varied terrains and obstacles (other rows). Because soft actor-critic learns robust policies, due to entropy maximization at training time, the policy can readily generalize to these perturbations without any additional learning. The robot is able to walk up and down a slope (first row), ram through an obstacle made of wooden blocks (second row), and step down stairs (third row) without difficulty, despite not being trained in these settings. To our knowledge, this experiment is the first example of a deep reinforcement learning algorithm learning underactuated quadrupedal locomotion directly in the real world without any simulation or pretraining. We have included videos of the the training process and evaluation on our project website\textsuperscript{\ref{footnote:webpage}}\!\!.

\subsection{Dexterous Hand Manipulation}
\begin{figure}[t]
    \centering
    \includegraphics[width=\textwidth]{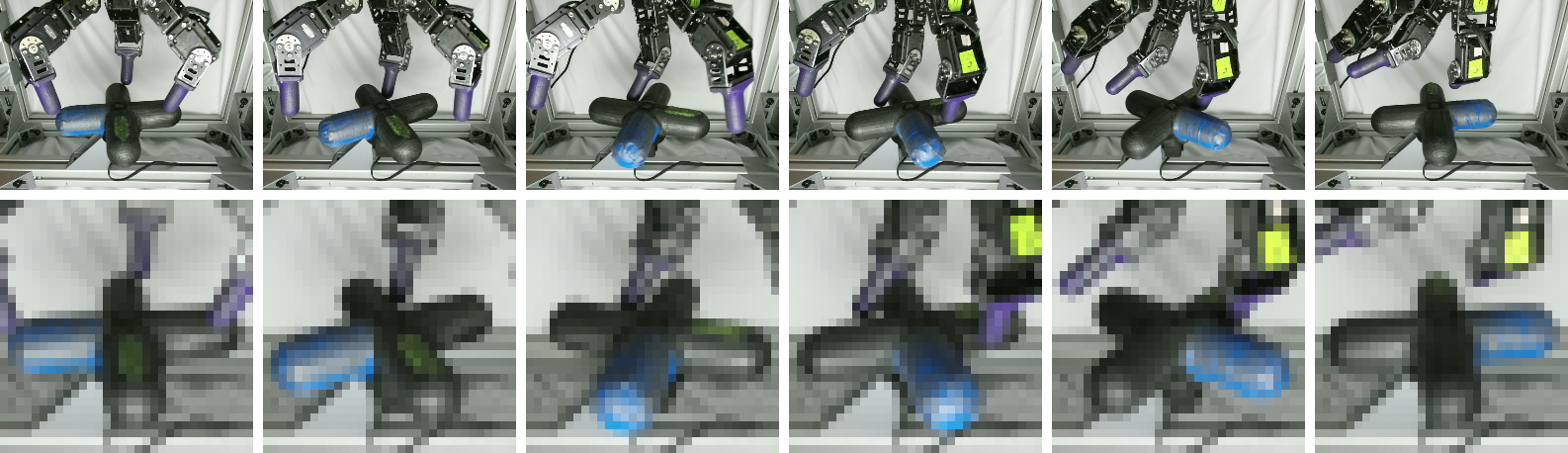}
    \caption{\small Valve rotation task with Dynamixel Claw. Top row shows a high resolution image sequence of a rollout with the policy. Bottom row shows the 32x32 pixel image observations of the same situations. The policy is also provided the joint positions and velocities of the fingers, but it has to infer the valve position from the image.}
	\label{fig:claw_from_pixels}
\end{figure}

Our second real-world robotic task involves training a 3-finger dexterous robotic hand to manipulate an object. The hand is based on the ``dclaw'' hand, discussed by \citep{zhu2018dexterous}. This hand has 9 DoFs, each controlled by a Dynamixel servo-motor. The policy controls the hand by sending target joint angle positions for the on-board PID controller. The manipulation task requires the hand to rotate a ``valve''-like object (resembling a sink faucet), as shown in \autoref{fig:claw_from_pixels}. In order to perceive the valve, the robot must use raw RGB images, which are illustrated in the second row of \autoref{fig:claw_from_pixels}. The images are processed in a neural network, consisting of two convolutional (four 3x3 filters) and max pool (3x3) layers, followed by two fully connected layers (256 units). The robot must rotate the valve into the correct position, with the colored part of the valve facing directly to the right, from any random starting position. The initial position of the valve is reset uniformly at random for each episode, forcing the policy to learn to use the raw RGB images to perceive the current valve orientation. A small motor is attached to the valve to automate resets and to provide the ground truth position for the determination of the reward function. The position of this motor is not provided to the policy.

This task is exceptionally challenging due to both the perception challenges and the physical difficulty of rotating the valve with such a complex robotic hand. As can be seen in the accompanying video on the project website\footnote{\href{https://sites.google.com/view/sac-and-applications/}{https://sites.google.com/view/sac-and-applications/}\label{footnote:webpage}}\!\!, rotating the valve requires a complex finger gait where the robot moves the fingers over the valve in a coordinated pattern, and stops precisely at the desired position.

\begin{figure}
    \centering
    \begin{subfigure}{0.32\textwidth}
        \includegraphics[width=\textwidth, trim={0 0 0 0}, clip]{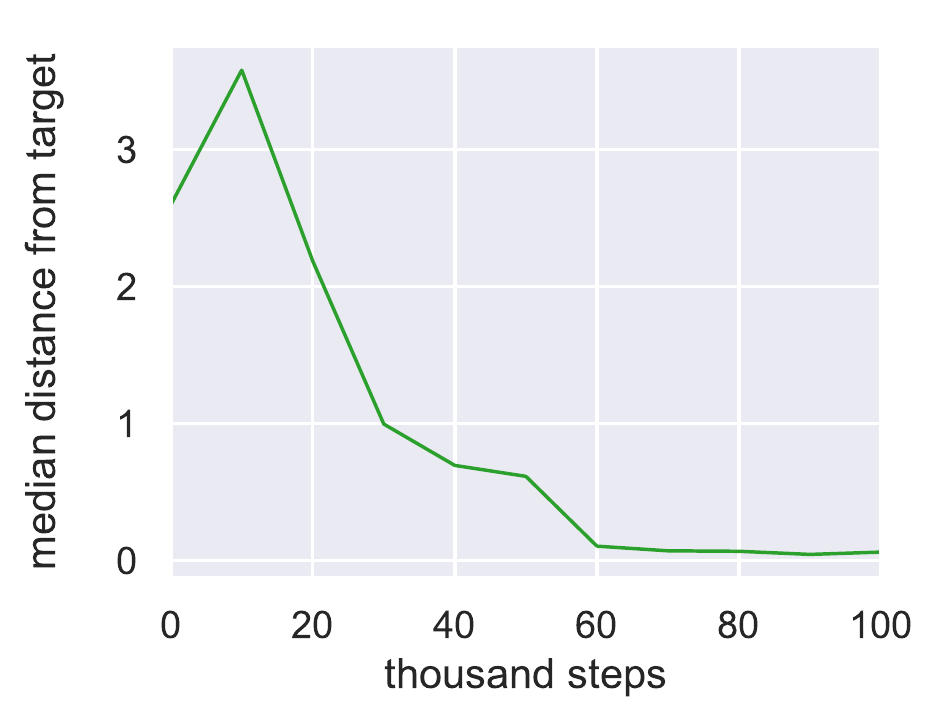}
        \caption{Learning from valve joint angle.}
        \label{fig:claw_non_vision}
    \end{subfigure}
    \hspace{10mm}
    \begin{subfigure}{0.32\textwidth}
        \includegraphics[width=\textwidth, trim={0 0 0 0}, clip]{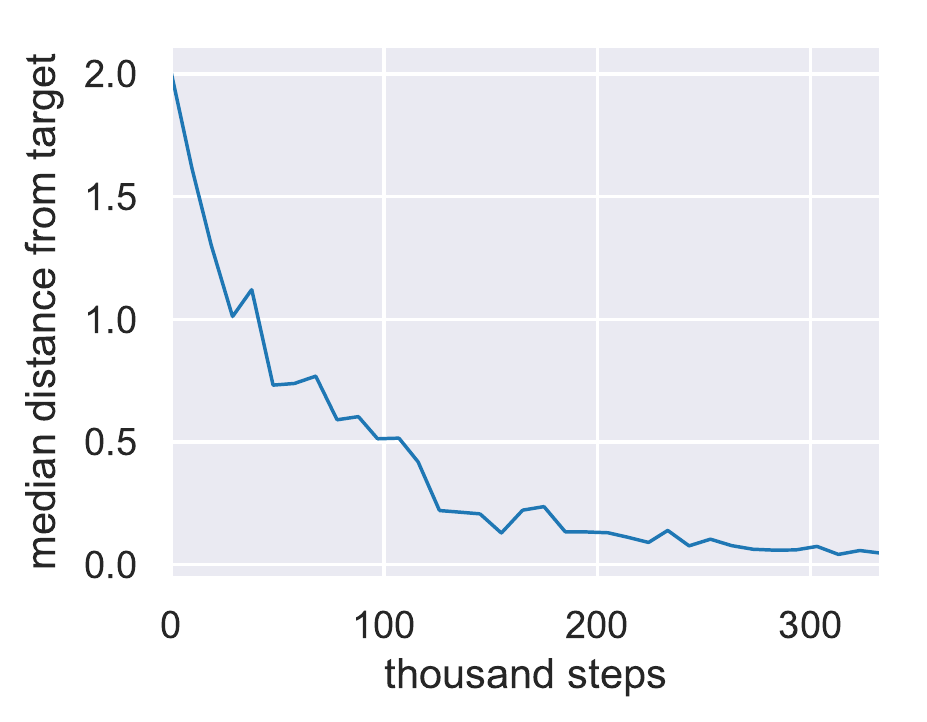}
        \caption{Learning from images.}
        \label{fig:claw_vision}
    \end{subfigure}
    
    \caption{\small Learning curves for the valve rotation task. The curves correspond to the median distance (measured in radians) of the valve from the target angle during a rollout. (a) Learning without images takes about 3 hours. In this case, the valve is reset to point away from the target initially. (b) Learning from images takes about 20 hours. The initial valve position is sampled uniformly at random.}
	\label{fig:claw_learning_curves}	
\end{figure}

Learning this task directly from raw RGB images requires 300k environment interaction steps, which is the equivalent of 20 hours of training, including all resets and neural network training time (\autoref{fig:claw_learning_curves}). To our knowledge, this task represents one of the most complex robotic manipulation tasks learned directly end-to-end from raw images in the real world with deep reinforcement learning, without any simulation or pretraining. We also learned the same task without images by feeding the valve position directly to the neural networks. In that case, learning takes 3 hours, which is substantially faster than what has been reported earlier on the same task using PPO (7.4 hours) \cite{zhu2018dexterous}.

\section{Conclusion}
In this article, we presented soft actor-critic (SAC), an off-policy maximum entropy deep reinforcement learning algorithm that provides sample-efficient learning while retaining the benefits of entropy maximization and stability. Our theoretical results derive soft policy iteration, which we show to converge to the optimal policy. From this result, we can formulate a practical soft actor-critic algorithm that can be used to train deep neural network policies, and we empirically show that it matches or exceeds the performance of state-of-the-art model-free deep RL methods, including the off-policy TD3 algorithm and the on-policy PPO algorithm without any environment specific hyperparameter tuning. Our real-world experiments indicate that soft actor-critic is robust and sample efficient enough for robotic tasks learned directly in the real world, such as locomotion and dexterous manipulation. To our knowledge, these results represent the first evaluation of deep reinforcement learning for real-world training of underactuated walking skills with a quadrupedal robot, as well as one of the most complex dexterous manipulation behaviors learned with deep reinforcement learning end-to-end from raw image observations.

\subsubsection*{Acknowledgments}
We would like to thank Vitchyr Pong and Haoran Tang for constructive discussions during the development of soft actor-critic, Vincent Vanhoucke for his support towards the project at Google, and Amazon for providing computing support.

\bibliography{references}
\bibliographystyle{icml2018}

\section*{Appendix}
\appendix
% \section{Maximum Entropy Objective}
\section{Infinite Horizon Discounted Maximum Entropy Objective}
\label{app:infinite_horizon_objective}
The exact definition of the discounted maximum entropy objective is complicated by the fact that, when using a discount factor for policy gradient methods, we typically do not discount the state distribution, only the rewards. In that sense, discounted policy gradients typically do not optimize the true discounted objective. Instead, they optimize average reward, with the discount serving to reduce variance, as discussed by \citet{thomas2014bias}. However, we can define the objective that \emph{is} optimized under a discount factor as
\begin{align}
J(\policy) = \sum_{t=0}^\infty \E{(\st,\at) \sim \rho_\policy}{ \sum_{l=t}^\infty \discount^{l-t} \E{\state_l\sim\pdyn,\action_l\sim\policy}{ \reward(\st,\at) + \alpha \ent(\policy(\voidarg|\st))|\st,\at}}.
\end{align}
This objective corresponds to maximizing the discounted expected reward and entropy for future states originating from every state-action tuple $(\st,\at)$ weighted by its probability $\rho_\pi$ under the current policy.
%%SL.12.25: some discussion of where this comes from? else it is a bit abstract. TH: added.

\section{Proofs}
%%SL.12.25: need to state the lemma before proving it, else readers have to flip back to learner in the paper all the time. TH: OK

\subsection{\autoref{lem:soft_policy_evaluation}}
\label{app:lem_soft_policy_evaluation}

\textbf{\autoref{lem:soft_policy_evaluation}} (Soft Policy Evaluation).\textit{
Consider the soft Bellman backup operator $\mathcal{T}^\policy$ in \autoref{eq:soft_bellman_backup_op} and a mapping $Q^0: \sspace \times \aspace\rightarrow \reals$ with $|\aspace|<\infty$, and define $\Q^{k+1} = \mathcal{T}^\policy \Q^k$. Then the sequence $Q^k$ will converge to the soft Q-value of $\policy$ as $k\rightarrow \infty$.}
\begin{proof}
Define the entropy augmented reward as $\reward_\policy(\st, \at) \triangleq \reward(\st, \at)  + \E{\stp\sim\pdyn}{\entropy\left(\policy(\voidarg|\stp)\right)}$ and rewrite the update rule as 
\begin{align}
\Q(\st, \at) \leftarrow \reward_\policy(\st, \at) + \discount \E{\stp\sim\pdyn,\atp \sim \policy}{Q(\stp, \atp)}
\end{align}
and apply the standard convergence results for policy evaluation~\citep{sutton1998reinforcement}. The assumption $|\aspace|<\infty$ is required to guarantee that the entropy augmented reward is bounded.
\end{proof}

\subsection{\autoref{lem:policy_improvement}}
\label{app:lem_policy_improvement}

\textbf{\autoref{lem:policy_improvement}} (Soft Policy Improvement).\textit{
Let $\policy_\mathrm{old} \in \Pi$ and let $\policy_\mathrm{new}$ be the optimizer of the minimization problem defined in \autoref{eq:constrainted_policy_fitting}. Then $\Q^{\policy_\mathrm{new}}(\st, \at) \geq \Q^{\policy_\mathrm{old}}(\st, \at)$ for all $(\st, \at) \in \sspace\times\aspace$ with $|\aspace|<\infty$.}
\begin{proof}
%We will drop the state and action arguments from the following derivation for improved readability. 
Let $\policyold\in \Pi$ and let $\Q^\policyold$ and $\V^\policyold$ be the corresponding soft state-action value and soft state value, and let $\policynew$ be defined as 
\begin{align}
\policynew(\voidarg|\st) &= \arg \min_{\policy' \in \Pi}\kl{\policy'(\voidarg|\st)}{\exp\left(Q^\policyold(\st,\voidarg) - \log Z^\policyold(\st)\right)}\notag\\
 &= \arg\min_{\policy'\in\Pi}J_\policyold(\policy'(\voidarg|\st)).
\end{align}
It must be the case that $J_\policyold(\policynew(\voidarg|\st)) \leq J_\policyold(\policyold(\voidarg|\st))$, since we can always choose $\policynew = \policyold\in\Pi$. Hence
\begin{align}
\resizebox{1\textwidth}{!}{$
\E{\at\sim\policynew}{\log \policynew(\at|\st) - Q^\policyold(\st, \at) + \log Z^\policyold(\st)}
\leq \E{\at\sim\policyold}{\log \policyold(\at|\st) - Q^\policyold(\st,\at) + \log Z^\policyold(\st)}$},
\end{align}
and since partition function $Z^\policyold$ depends only on the state, the inequality reduces to
\begin{align}
\E{\at\sim\policynew}{Q^\policyold(\st, \at) - \log \policynew(\at|\st)} \geq V^\policyold(\st).
\label{eq:soft_value_bound}
\end{align}
Next, consider the soft Bellman equation:
\begin{align}
Q^\policyold(\st, \at) &= \reward(\st, \at) + \discount\E{\stp\sim\pdyn}{V^\policyold(\stp)}\notag\\
&\leq \reward(\st, \at) + \discount\E{\stp\sim\pdyn}{\E{\atp\sim\policynew}{Q^\policyold(\stp, \atp) - \log \policynew(\atp|\stp)}}\notag\\
&\ \  \vdots\notag\\
& \leq Q^\policynew(\st, \at),
\end{align}
where we have repeatedly expanded $Q^\policyold$ on the RHS by applying the soft Bellman equation and the bound in \autoref{eq:soft_value_bound}. Convergence to $\Q^\policynew$ follows from \autoref{lem:soft_policy_evaluation}.
\end{proof}

\subsection{\autoref{the:soft_policy_iteration}}
\label{app:the_soft_policy_iteration}

\textbf{\autoref{the:soft_policy_iteration}} (Soft Policy Iteration). \textit{
Repeated application of soft policy evaluation and soft policy improvement to any $\policy\in\Pi$ converges to a policy $\policy\opt$ such that $Q^{\policy\opt}(\st, \at) \geq Q^{\policy}(\st, \at)$ for all $\policy \in \Pi$ and $(\st, \at) \in \sspace\times\aspace$, assuming $|\aspace|<\infty$.}
\begin{proof}
Let $\policy_i$ be the policy at iteration $i$. By \autoref{lem:policy_improvement}, the sequence $Q^{\policy_i}$ is monotonically increasing. Since $Q^\policy$ is bounded above for $\policy \in \Pi$ (both the reward and entropy are bounded), the sequence converges to some $\policy\opt$. We will still need to show that $\policy\opt$ is indeed optimal. At convergence, it must be case that $J_\policyopt(\policyopt(\voidarg|\st)) < J_\policyopt(\policy(\voidarg|\st))$ for all $\policy\in\Pi$, $\policy\neq \policy\opt$. Using the same iterative argument as in the proof of \autoref{lem:policy_improvement}, we get $Q^\policyopt(\st, \at) > Q^\policy(\st, \at)$ for all $(\st, \at)\in \sspace\times\aspace$, that is, the soft value of any other policy in $\Pi$ is lower than that of the converged policy. Hence $\policyopt$ is optimal in $\Pi$.
\end{proof}

\section{Enforcing Action Bounds}
\label{app:action_bounds}
%%SL.12.25: I don't think this appendix is referenced anywhere, can you reference it? TH: done.

%%SL.12.25: generally, technical description of the method should use present tense, rather than past tense TH: good point. fixed.
We use an unbounded Gaussian as the action distribution. However, in practice, the actions needs to be bounded to a finite interval. To that end, we
apply an invertible squashing function 
($\tanh$) to the Gaussian samples, and employ the change of variables formula to compute the likelihoods of the bounded actions. In the other words, let $\urv\in\reals^D$ be a random variable and $\mu(\urv|\state)$ the corresponding density with infinite support. Then $\action = \tanh(\urv)$, where $\tanh$ is applied elementwise, is a random variable with support in $(-1, 1)$ with a density given by
\begin{align}
\policy (\action|\state) &= \mu(\urv|\state)\left|\det \left(\frac{\mathrm{d}\action}{\mathrm{d}\urv} \right)\right|^{-1}.
\end{align}
Since the Jacobian $\nicefrac{\mathrm{d}\action}{\mathrm{d}\urv} = \mathrm{diag}(1 - \tanh^2(\urv))$ is diagonal, the log-likelihood has a simple form 
%%SL.12.25: I don't understand, are these two equations supposed to be equivalent? is that obvious? it doesn't seem obvious to me TH: added explanation
\begin{align}
\log\policy (\action|\state) &= \log \mu(\urv|\state) - \sum_{i=1}^D\log\left(1 - \tanh^2(u_i)\right),
\end{align}
where $u_i$ is the $i^\mathrm{th}$ element of $\urv$.

\newpage
\section{Hyperparameters}
\label{app:hypers}
\autoref{tab:shared_params} lists the common SAC parameters used in the comparative evaluation in \autoref{fig:training_curves}.
\begin{table}[H]
\renewcommand{\arraystretch}{1.1}
\centering
\caption{SAC Hyperparameters}
\label{tab:shared_params}
\vspace{1mm}
\begin{tabular}{l| l }
\toprule
Parameter &  Value\\
\midrule
optimizer &Adam \citep{kingma2014adam}\\
learning rate & $3 \cdot 10^{-4}$\\
discount ($\discount$) &  0.99\\
replay buffer size & $10^6$\\
number of hidden layers (all networks) & 2\\
number of hidden units per layer & 256\\
number of samples per minibatch & 256\\
entropy target & $-\dim\left(\aspace\right)$ (\eg, -6 for HalfCheetah-v1)\\
nonlinearity & ReLU\\
target smoothing coefficient ($\tau$)& 0.005\\
target update interval & 1\\
gradient steps & 1\\
\bottomrule
\end{tabular}
\end{table}

\end{document}